\documentclass{article}

\usepackage{microtype}
\usepackage{subfigure}
\usepackage{booktabs} 

\usepackage[accepted]{icml2023}

\usepackage{graphicx}
\usepackage{amsmath}
\usepackage{amssymb}
\usepackage{colortbl}
\usepackage{xcolor}
\usepackage[pagebackref,breaklinks,colorlinks]{hyperref}

\usepackage[capitalize,noabbrev]{cleveref}
\crefname{section}{Sec.}{Secs.}
\Crefname{section}{Section}{Sections}
\Crefname{table}{Table}{Tables}
\crefname{table}{Tab.}{Tabs.}
\newcommand{\eg}{e.g.,}

\icmltitlerunning{X-Paste: Revisiting Scalable Copy-Paste for Instance Segmentation using CLIP and StableDiffusion}

\begin{document}

\twocolumn[
\icmltitle{X-Paste: Revisiting Scalable Copy-Paste for Instance Segmentation \\using CLIP and StableDiffusion}



\icmlsetsymbol{equal}{*}

\begin{icmlauthorlist}
\icmlauthor{Hanqing Zhao}{equal,ustc}
\icmlauthor{Dianmo Sheng}{equal,ustc}
\icmlauthor{Jianmin Bao}{ms}
\icmlauthor{Dongdong Chen}{ms}
\icmlauthor{Dong Chen}{ms}
\icmlauthor{Fang Wen}{ms}
\icmlauthor{Lu Yuan}{ms}
\icmlauthor{Ce Liu}{ms}
\icmlauthor{Wenbo Zhou}{ustc}
\icmlauthor{Qi Chu}{ustc}
\icmlauthor{Weiming Zhang}{ustc}
\icmlauthor{Nenghai Yu}{ustc}
\end{icmlauthorlist}

\icmlaffiliation{ustc}{University of Science and Technology of China}
\icmlaffiliation{ms}{Microsoft}

\icmlcorrespondingauthor{Jianmin Bao}{jianmin.bao@microsoft.com}
\icmlcorrespondingauthor{Wenbo Zhou}{welbeckz@ustc.edu.cn}

\icmlkeywords{Instance Segmentation, Copy Paste Augmentation}

\vskip 0.3in
]



\printAffiliationsAndNotice{\icmlEqualContribution} 

\begin{abstract}
Copy-Paste is a simple and effective data augmentation strategy for instance segmentation. By randomly pasting object instances onto new background images, it creates new training data for free and significantly boosts the segmentation performance, especially for rare object categories. Although diverse, high-quality object instances used in Copy-Paste result in more performance gain, previous works utilize object instances either from human-annotated instance segmentation datasets or rendered from 3D object models, and both approaches are too expensive to scale up to obtain good diversity. In this paper, we revisit Copy-Paste at scale with the power of newly emerged zero-shot recognition models (e.g., CLIP) and text2image models (e.g., StableDiffusion). We demonstrate for the first time that using a text2image model to generate images or zero-shot recognition model to filter noisily crawled images for different object categories is a feasible way to make Copy-Paste truly scalable. To make such success happen, we design a data acquisition and processing framework, dubbed ``X-Paste", upon which a systematic study is conducted. On the LVIS dataset, X-Paste provides impressive improvements over the strong baseline CenterNet2 with Swin-L as the backbone. Specifically, it archives \textbf{+2.6} box AP and \textbf{+2.1} mask AP gains on all classes and even more significant gains with  \textbf{+6.8} box AP \textbf{+6.5} mask AP on long-tail classes. Our code and models are available at \href{https://github.com/yoctta/XPaste}{https://github.com/yoctta/XPaste}.
\end{abstract}

\section{Introduction}
\label{sec:intro}

Instance segmentation \cite{dai2016instance,he2017mask,hafiz2020survey} is a fundamental task in computer vision with very broad applications. In order to get plausible performance for one specific category, most existing methods~\cite{li2022mask,dong2022cswin,Swin,he2017mask} rely on a large number of images annotated for this category, which is not only expensive but also time-consuming. This also makes expanding the object category coverage extremely hard. In fact, the real-world images often follow a long-tail distribution, so collecting enough images for some rare categories itself is already very difficult. Therefore, it is of great value to study how to augment and create training data in an efficient and scalable way.

As one simple yet effective data augmentation strategy, Copy-Paste \cite{Copy-Paste,Simple-Copy-Paste,Contextual-Copy-Paste} has been extensively studied to improve data efficiency. By randomly pasting object instances onto background images, it can generate a combinatorial number of training data for free and boost the instance segmentation model performance, especially for rare categories. Intuitively, if we can utilize more diverse object instances in Copy-Paste, more performance gain can be achieved. However, the object instances used in existing methods are either from the instance segmentation dataset itself \cite{Simple-Copy-Paste,Contextual-Copy-Paste} or rendered from external 3D models ~\cite{Copy-Paste}. In this paper, we argue that both these two manners are not scalable and have not exploited the full potential of Copy-Paste. More specifically, for the former manner, on one hand, collecting human-annotated data is not scalable. On the other hand, the limited instance number of rare categories also results in limited diversity of Copy-Paste data. In the latter manner, building/collecting 3D models itself is super difficult and not scalable.    

In this paper, we propose a new object instance acquisition and processing framework, dubbed ``X-Paste". X-Paste is built upon Copy-Paste~\cite{Simple-Copy-Paste} to train the instance segmentation model but aims to make Copy-Paste more scalable, i.e., \emph{obtain large-scale object instances with high-quality masks (``oxygen of Copy-Paste") for unlimited categories in an efficient and automatic way}. X-Paste consists of four major modules: 
Object Instance Acquisition, Instance Mask Generation, Instance Filtering, and Instance Composition with Background Images.

To get large-scale images for different categories in a scalable way, the core idea of the Object Instance Acquisition module is to take full advantage of the newly emerged powerful zero-shot text2image generative model like DALLE~\cite{dalle2} and StableDiffusion\cite{ldm} (used because of the accessibility),  or zero-shot recognition model like CLIP ~\cite{CLIP} and Florence \cite{yuan2021florence}, both of which are trained on web-scale image-text pairs. For the text2image generative models, by feeding different text prompts of one specific category, they can generate very diverse images with different appearances, viewpoints and styles. In contrast, previous methods~\cite{Copy-Paste, Simple-Copy-Paste, Contextual-Copy-Paste} ``Copying" images from training datasets or rendering from limited 3D models suffer worse diversity. For the zero-shot recognition model, even though it cannot generate images like text-to-image model, it enables us to filter high-quality data for any category from large-scale web-retrieved images, which is unachievable for previous close-set (limited category coverage) image recognition models. Moreover, our study shows that these two data acquisition methods can be combined together to get better performance.

Given the generated or retrieved images, the following Instance Mask Generation module is designed to get the instance masks, which are required in Copy-Paste for composition. We observe that the generated or web-retrieved images for different categories are usually object-centric. Especially for generated images, they are often with simple background. So they are relatively easier to be segmented with off-the-shelf salient object segmentation models~\cite{U2Net, selfreformer,UFO}. One advantage of these models is that  they can estimate the mask for any category, i.e., category-agnostic, making it possible for generating instances under open-vocabulary setting. Note that, in order to get precise segmentation mask for web-retrieved images, we conduct the background analysis and remove the images with complex backgrounds.

After getting the instance masks, X-Paste performs Instance Filtering to remove samples with wrong categories or imprecise mask segmentation. We further utilize the CLIP model to calculate the similarity between the segmented results and the given category, and then remove samples with low similarity. After that, we can have a great number of object instances for different categories.  
Finally, the Instance Composition module generates the training data to train the segmentation model by pasting the object instances onto different background images. We study various composition strategies and find that assigning a random scale and location for these instances can get significant performance boost. We also find that advanced blending methods like Poisson blending show no more gains.

Our X-Paste presents an important attempt that utilizes the high-quality zero-shot text2image model and recognition model to help image understanding tasks, even though it may look very intuitive and straightforward. Its great scalability can help make large or even open vocabulary instance segmentation happen, by alleviating the pressure of human annotation for instance segmentation. It can act as one "plug and play" component for any instance segmentation framework without any architecture change or inference overhead. To the best of our knowledge, this is the first work showing how to make Copy-Paste at scale and further explore its potential.

We perform extensive experiments to validate the superiority of X-Paste, On COCO dataset, X-Paste achieves 57.0 box mAP and 48.6 mask mAP, outperforming baseline by +1.7 mAP and +0.9 mAP. On LVIS dataset, we achieve 44.4 mask mAP for all objects and 43.3 mask mAP for rare objects. yield +2.1 and +6.5 gains over baseline. Our method can also benefit open-vocabulary setting, our model achieves 31.8 mask mAP for all objects and 21.4 mask mAP for rare objects, +1.6, and +5.0 mAP over baseline.
Meanwhile, we perform a comprehensive analysis of our framework X-Paste the demonstrate the effectiveness of each step.

\section{Related Works}
\label{sec:relatedwork}

\noindent\textbf{Instance Detection and Segmentation.}
Instance detection and segmentation \cite{he2017mask,dai2021dynamic,meng2022detection,li2022mask} has been extensively studied in the past decades. Given an image, instance detection is to locate the exact position of object instance and determine the category of the instance, while instance segmentation further needs to get the fine-grained mask of the instance. Since Mask-RCNN \cite{he2017mask}, these two tasks are usually integrated into an unified two-stage framework, so we use instance segmentation to denote these two tasks. Recently, long-tail and open-vocabulary instance segmentation are drawing more attention, since they are more realistic settings in real applications.

LVIS dataset \cite{LVIS} is one of the most challenging long-tail instance segmentation dataset. 
Due to the high imbalance and limited instances, current instance segmentation models often struggle to handle such rare categories well. Therefore, different techniques have been proposed, e.g., data re-sampling and loss re-weighting\cite{wang2020devil,mahajan2018exploring,zang2021fasa,tan2021equalization,wang2021seesaw}, score normalization \cite{pan2021model}, and data augmentation \cite{Simple-Copy-Paste}. In this paper, we focus on Copy-Paste based augmentation and 
study how to make it scalable to further explore its potential. In this sense, our study is complementary to other non-Copy-Paste techniques.

Open-vocabulary instance detection (OVD) aims to detect target/novel class (unseen) objects not present in the training/base class (seen) vocabulary at inference. Therefore, it can be viewed as a hard corner of long-tail instance detection, where no instances exist for some categories. To improve the open-vocabulary performance, some existing methods leverage the large-scale pretrained vision-language models (e.g., CLIP) to transfer the rich knowledge into the classifier, including OVR-CNN \cite{ovr-cnn}, ViLD \cite{ViLD}, OpenSeg \cite{OpenSeg}, RegionCLIP \cite{RegionCLIP}, DetPro \cite{DetPro} and PromptDet \cite{PromptDet}. And the latest work Detic \cite{Detic} proposes to utilize image classification data (i.e., ImageNet-22k\cite{deng2009imagenet}) to help train the classifier via image-level supervision. In the following experiments, we will show our X-Paste, as a simple ``Plug and Play" augmentation strategy without the need of any model architecture or training strategy change, can already enable the vanilla instance segmentation model plausible open-vocabulary capability. Again, it is complementary to the aforementioned open vocabulary techniques.

\noindent\textbf{Data Boosting for Instance Segmentation.}
Since most existing instance segmentation models are data hungry, a lot of efforts have been devoted to improve the performance from the data boosting perspective. Based on the boosted data type, they can be roughly divided into synthesis-based and retrieval-based.

For synthesis-based methods, Early works use graphics based renderings\cite{hinterstoisser2018pre,su2015render, hodavn2019photorealistic} or computer games\cite{richter2016playing,richter2017playing} to generate high quality labelled data. However, these methods often suffer from the huge domain gap between real and pure synthetic data.
Then composing real images are introduced in following works, including Copy-Paste \cite{Copy-Paste}, Contextual Copy-Paste \cite{Contextual-Copy-Paste}, and Instaboost \cite{Instaboost}. And Simple Copy-Paste \cite{Simple-Copy-Paste} further indicates that simply pasting real segmented objects randomly onto background images already works very well, without the need of advanced strategies like contextual modeling. The main problem of existing Copy-Paste methods is that they are not scalable and suffer from limited instance diversity, since they get the object instances either from the instance dataset itself or rendered from 3D models. 

Retrieval is another way to collect large-scale data. with the development of large-scale image-text dataset LAION\cite{laion} and image classification dataset ImageNet-22k\cite{deng2009imagenet}, some approaches \cite{PromptDet,Detic,hong2017weakly,jin2017webly,shen2018bootstrapping,wei2016stc} study how to leverage such data to help instance segmentation. But since these data themselves do not have the precise instance annotation, some dedicated designs are needed, e.g., pseudo labeling \cite{RegionCLIP} or freezing the location branch and only train the classifier with the image-level supervision in \cite{Detic}.

The key difference between our X-Paste and the above efforts is that we try to utilize the text2image model synthesized data and retrieval data under the same simple Copy-Paste framework and make it scalable, without the need of any algorithm change. Moreover, it is orthogonal to the efforts like leveraging the pretrained CLIP model for knowledge distillation or image-level supervision. 

\noindent \textbf{Zero-shot Recognition and Text2Image Generation.}
Vision-Language Pre-training (VLP) has recently made very encouraging breakthrough. By conducting contrastive learning upon web-scale image-text data, the representative works, including CLIP\cite{CLIP}, MaskCLIP \cite{dong2022maskclip}, Florence \cite{yuan2021florence}, ALIGN\cite{ALIGN} and OmniVL\cite{wang2022omnivl}, have shown great power in aligning the visual feature with textual feature, enabling the real zero-shot recognition rather than traditional close-set recognition. Also pretraining on the web-scale image-text data, another breakthrough is about zero-shot text2image generative models. Representative works include DALL-E \cite{dalle,glide,dalle2}, CogView \cite{cogview}, Parti \cite{parti}, Imagen \cite{imagen}, VQ-Diffusion \cite{gu2022vector}, StableDiffusion\cite{ldm} and Frido\cite{fan2023frido}. By feeding free-form text, these models can generate high-fidelity images that match the text input condition. Intuitively, such zero-shot text2image models are natural data generators, but very few prior works have used them to help image understanding training. In this paper, we present the first attempt that utilizes the powerful zero-shot recognition and text2image models to revisit Copy-Paste at scale and help instance segmentation.

\section{X-Paste}
\label{sec:method}

\begin{figure*}
   \centering
   \includegraphics[width=0.9\linewidth]{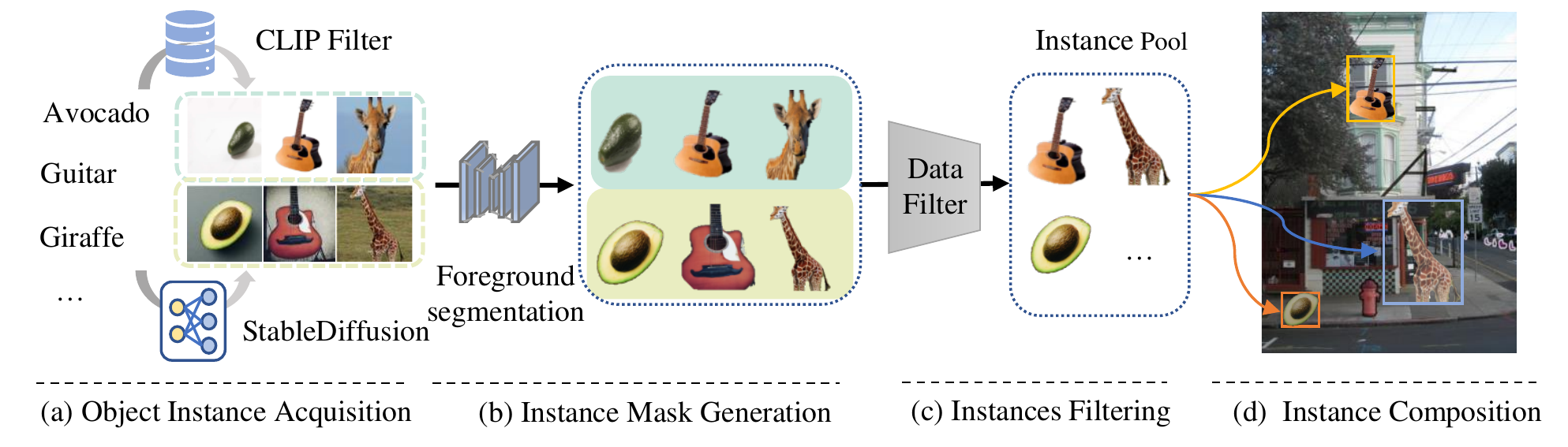}
   \vspace{-1em}
   \caption{Overview of the X-Paste pipeline. X-Paste first obtains filtered real images from CLIP and generated images from StableDiffusion. Then Instance Mask Generation module gets the pseudo mask for all images. After that, the Instance Filtering module removes these samples with inaccurate mask segmentation. Finally, we paste these resulting instances into a background image for training.}
   \label{fig:instance}
 \end{figure*}
 
\begin{figure}[h]
\begin{center}
\resizebox{0.95\linewidth}{!}{
\begin{tabular}{ccccc}
& \small{SRF} & \small{CLIPseg} & \small{UFO} & \small{U2Net}
\\
\includegraphics[width=1.4cm]{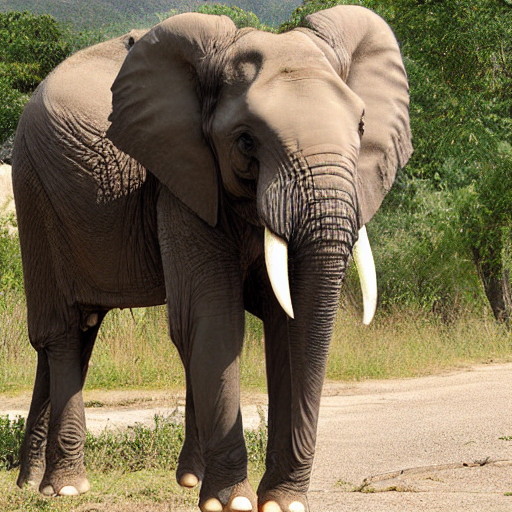}
&\includegraphics[width=1.4cm]{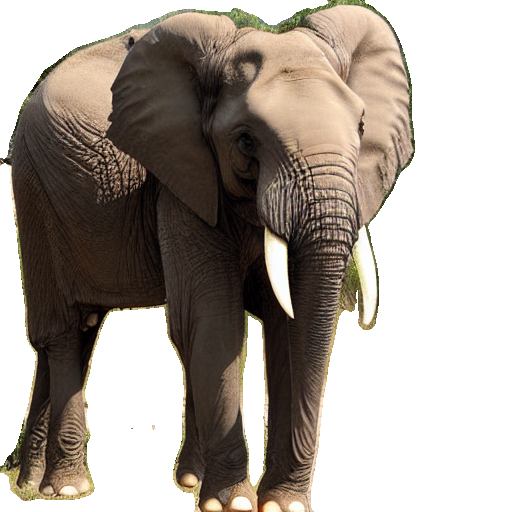}
&\includegraphics[width=1.4cm]{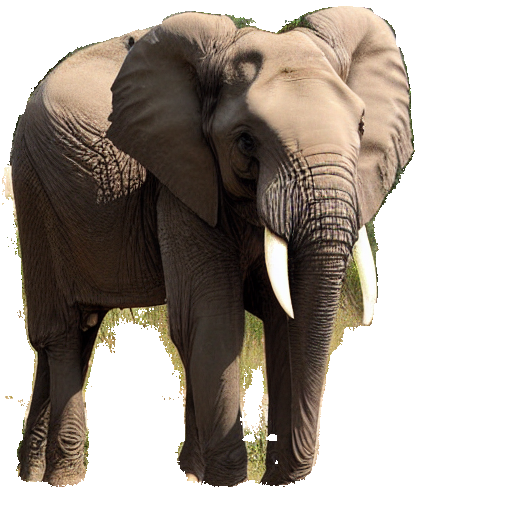}
&\includegraphics[width=1.4cm]{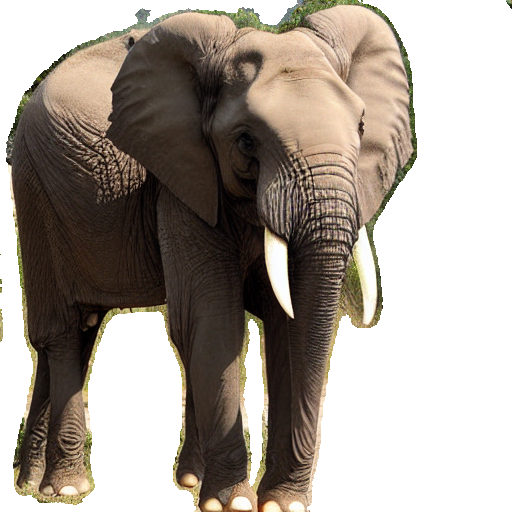}
&\includegraphics[width=1.4cm]{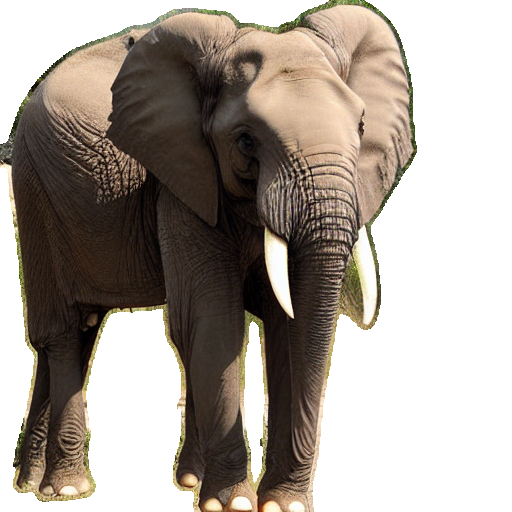}
\\
\small{CLIP score} & 0.2448 & 0.2328 & 0.2375 & 0.2451 \\
\includegraphics[width=1.4cm]{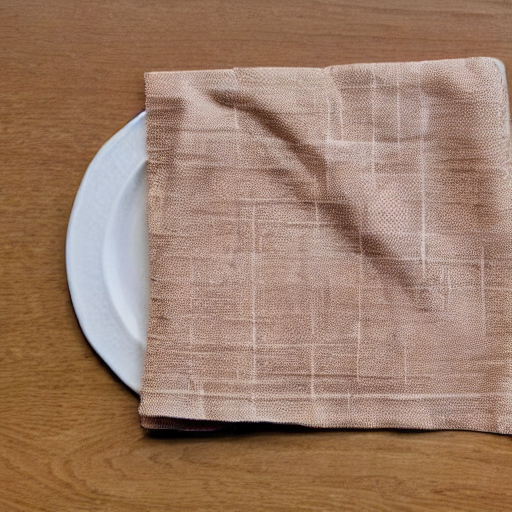}
&\includegraphics[width=1.4cm]{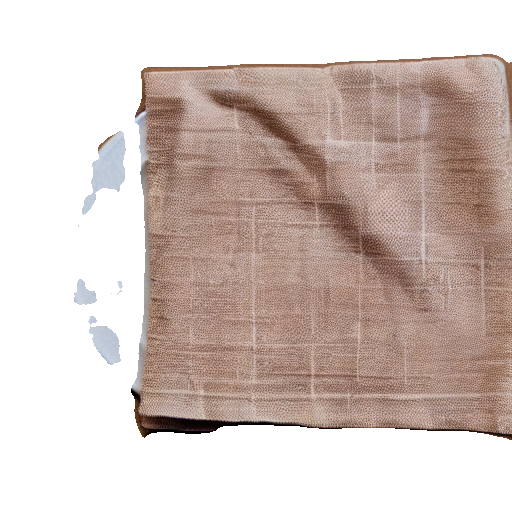}
&\includegraphics[width=1.4cm]{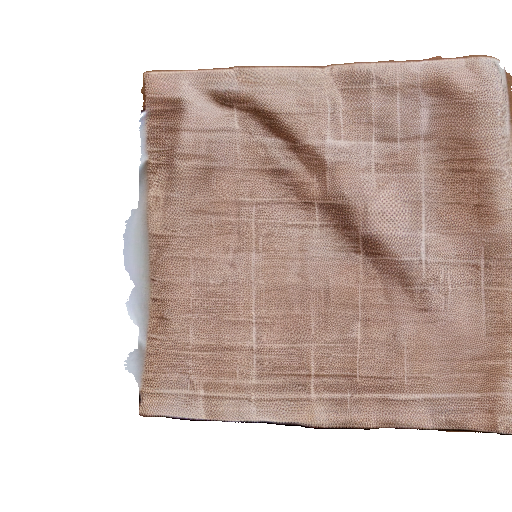}
&\includegraphics[width=1.4cm]{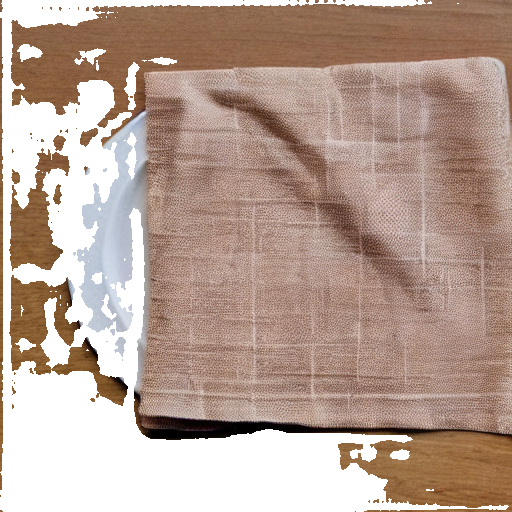}
&\includegraphics[width=1.4cm]{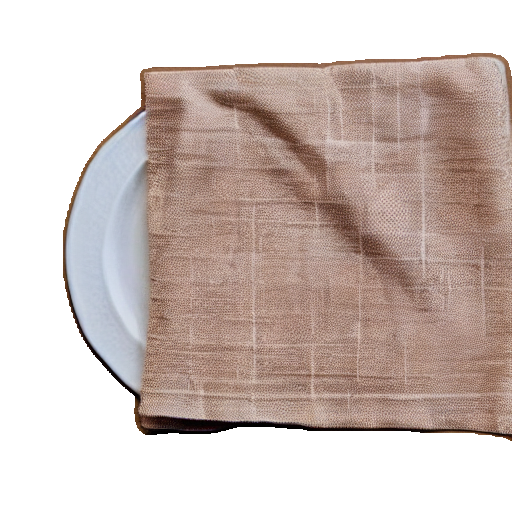}
\\
\small{CLIP score} & 0.2231 & 0.2425 & 0.2128 & 0.2301 \\
\includegraphics[width=1.4cm]{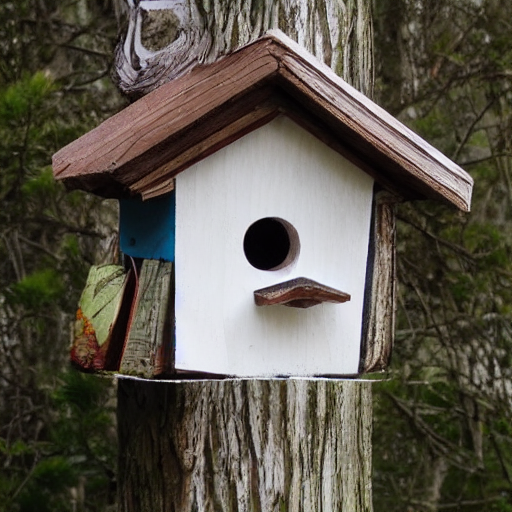}
&\includegraphics[width=1.4cm]{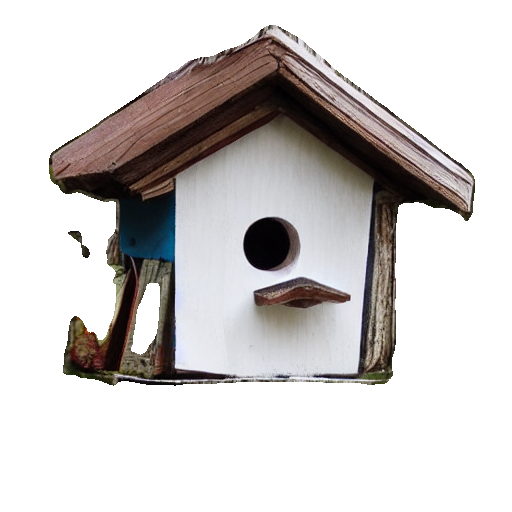}
&\includegraphics[width=1.4cm]{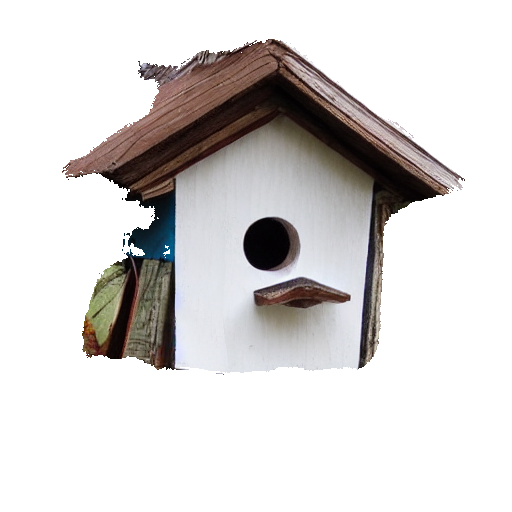}
&\includegraphics[width=1.4cm]{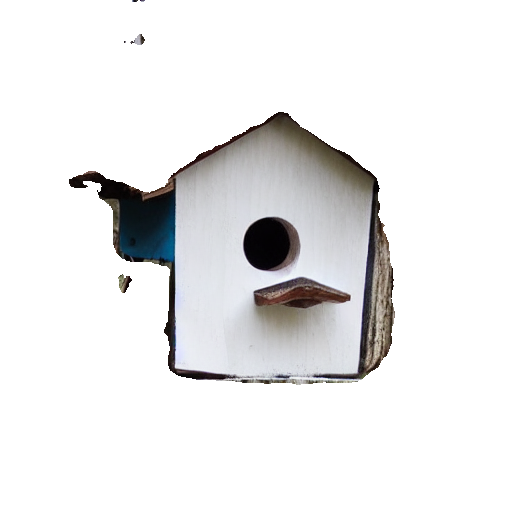}
&\includegraphics[width=1.4cm]{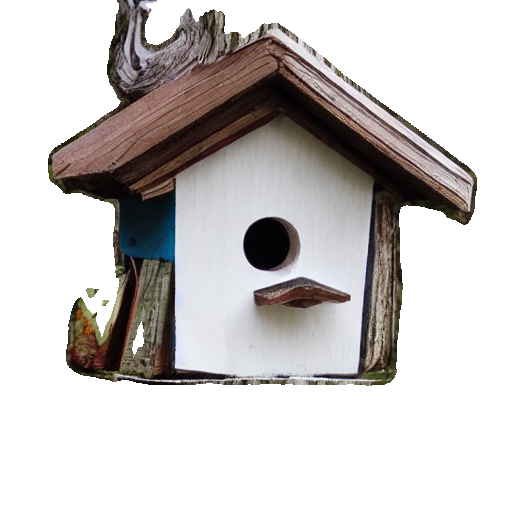}
\\
\small{CLIP score} & 0.2767 & 0.2776 & 0.1698 & 0.2760 \\
\end{tabular}
}
\end{center}
\caption{Predicted foregrounds of four segmentation methods and their CLIP score with the given text.}
\vspace{-1em}
\label{fig:seg}
\end{figure}
 
As described above, previous Copy-Paste methods~\cite{Copy-Paste, Simple-Copy-Paste, Contextual-Copy-Paste} utilize instances from internal instance segmentation dataset itself or rendered from 3D CG models, making them hard to scale up and exploit the full potentials of Copy-Paste. In this paper, we study how to make Copy-Paste scalable by acquiring external object instances in an automatic and efficient way. We propose to leverage generated images from zero-shot text2image model(\i.e., StableDiffusion~\cite{ldm}) or web-retrieved images filtered by zero-shot image classification model(\eg, CLIP~\cite{CLIP}). In order to use such images under the same simple Copy-Paste framework, we propose a complete data acquisition and processing framework ``X-Paste". As shown in Figure~\ref{fig:instance}, the overall framework is very simple and straightforward, consisting of four major modules: Object Instance Acquisition, Instance Mask Generation, Instance Filtering, and Instance Composition. In the following sections, we will elaborate each module in detail.

\subsection{Object Instance Acquisition}
\label{sec:method1}
To collect images of a specific category with diverse appearances, viewpoints, and styles, we provide two efficient and effective solutions based on recently emerged large-scale vision-language models (i.e., text2image model and image-text contrastive model), both of which show strong zero-shot capability and make data acquisition for any category scalable. In detail, as the first solution, we propose to directly apply the powerful zero-shot text2image generative model~\cite{ldm} to generate images for the interested categories. We obtain images of a specific category by feeding a text prompt like ``a photo of a single [\textit{Category Name}]". The word ``single" here is to encourage the model to generate a single instance in the image since the generative model may generate multiple instances for many categories. We find that these generated images are photorealistic, super diverse, and semantically match with the given category, as such text2image models are trained with web-scale data and show great combinatorial capability. In this paper, we use StableDiffusion V1.4 with PLMS sampler by default, as it is publicly open-sourced.

As the second scalable solution, we propose to crawl real images from the internet as the object instance source. Since web-crawled images are often very noisy, directly using all of them without filtering will destroy the model. Before the emergence of latest zero-shot image-text contrastive models, filtering images for different categories is not a trivial task, since most recognition models are trained on close-set categories. To make such recognition models perform well for any category, large-scale category-specific data needs to be collected, thus making it a chicken-egg problem. By conducting contrastive learning on web-scale image-text data, the latest vision-language alignment models like CLIP \cite{CLIP}, Florence \cite{yuan2021florence} make zero-shot image-text similarity matching possible. Therefore, we apply the CLIP model to calculate the semantic similarity between each category and the corresponding crawled images, and only keep the images with high semantic similarity. Moreover, to help the following Instance Mask Generation module generate high-quality instance masks, we further conduct the background analysis and only keep the images with simple backgrounds. In detail, we simply calculate the color histogram of each image and find the dominant color. Then we only keep the images in which over 40\% image pixels are close to the dominant color (color difference smaller than 5).

\subsection{Instance Mask Generation}
\label{sec:instance_mask_generation}
Since Copy-Paste needs the instance mask during composition, given the generated or crawled images, another challenge is to get the precise object instance masks. By going through the acquired images, we observe most of them are object-centric. This motivates us to use the class-agnostic foreground segmentation models to generate the instance mask. In detail, we have investigated three types of foreground segmentation algorithms: salient object segmentation, co-saliency segmentation and text-guided segmentation. We use prevalent methods U2Net \cite{U2Net} and SelfReformer (abbreviated as SRF) \cite{selfreformer} for salient object segmentation, UFO \cite{UFO} for co-saliency segmentation with a batch of images from each category, and CLIPseg \cite{CLIPseg} to segment the object with \textit{Category Name} as the text prompt. Since CLIPseg \cite{CLIPseg} produces very coarse segment masks, we convert the generated masks into the tri-map and use an image matting method \cite{matteformer} to further refine the mask. 

We observe all these segmentation methods will fail in some cases, as shown in Figure \ref{fig:seg}. So we propose a CLIP-guided selection strategy to select one instance mask as the pseudo annotation from these four methods for each image. Specifically, we calculate the cross-modal similarity between the image of segmented object in blank background and its category name. Our motivation is that better segmentation will get higher semantic similarity. To illustrate this motivation, we visualize the CLIP score in Figure \ref{fig:seg}.

\subsection{Instance Filtering}
\begin{figure}
   \centering
   \includegraphics[width=0.8\linewidth]{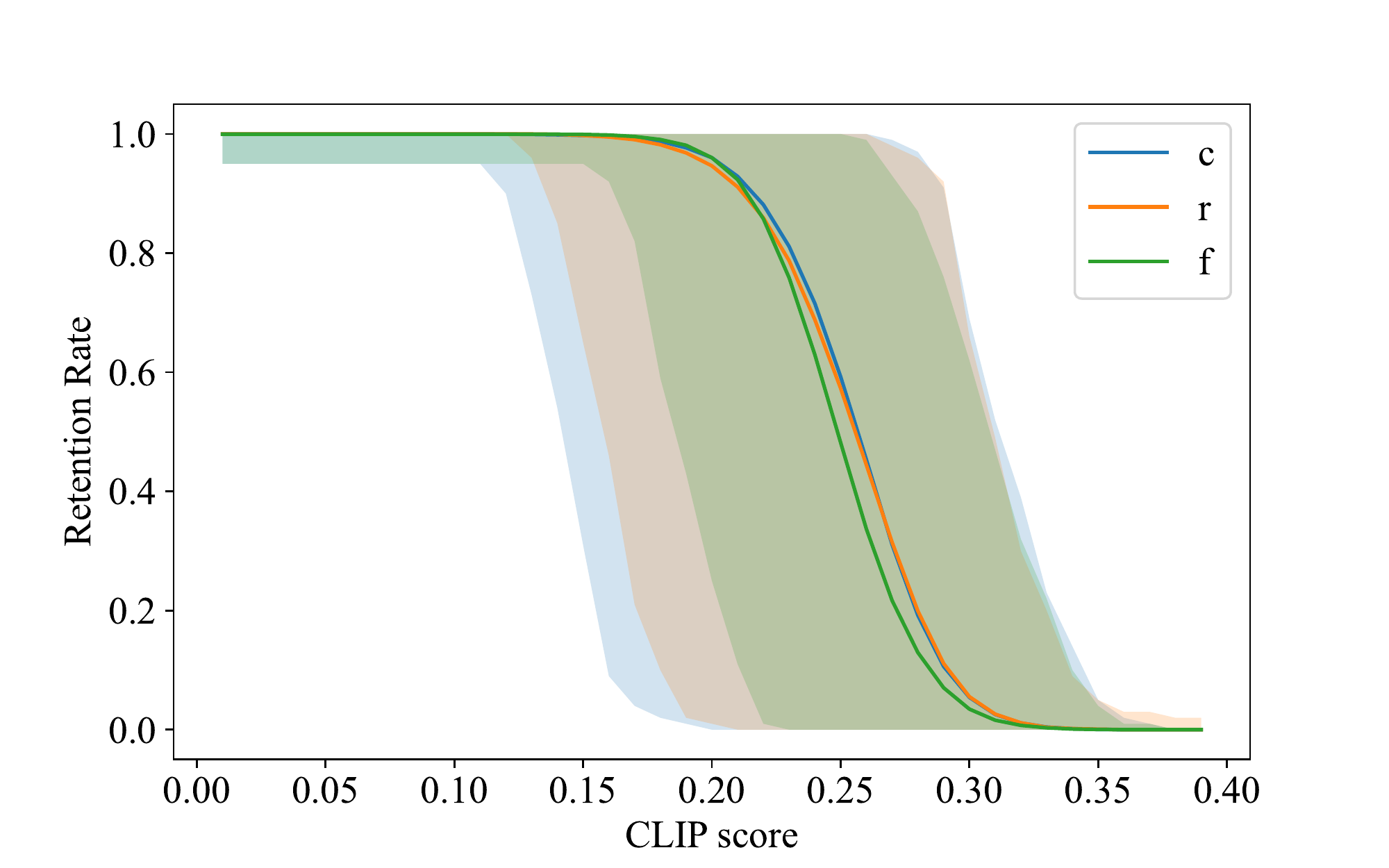}
   \caption{We show the retention rate of instances generated by StableDiffusion filter varying with CLIP score threshold. `r', `c', and `f' indicates ``rare,``common" and ``frequent" categories in LVIS dataset, the shadow areas represent the upper and lower bound of per-category CLIP score distribution, the lines represent entire CLIP score distribution.}
   \vspace{-1em}
   \label{fig:stable_CLIP}
 \end{figure}

To further remove object instances with low-quality masks, we apply several strategies in the Instance Filtering module. First, we filter out masks with areas less than 5\% or over 95\% of the whole image because these instances are highly possibly segmented incorrectly. Second, we select the instances with high semantic relevance to the category with pre-computed CLIP scores. Considering the different sensitivity of CLIP score for each category, we set the category-specific threshold $thres_i=min(t,max(\mathbf{C}_i)-d)$, where $t$ is the predefined CLIP score threshold, $d$ is a subtractive threshold for those classes with low image-text similarity, default as 0.01, $\mathbf{C}_i$ is the CLIP score set of instances from category $i$. Since Copy-Paste is especially helpful for rare categories in a long-tail dataset, it is more important to generate enough diverse instances of rare categories. Therefore, we are very curious whether generative models can generate high-quality instances for such rare categories. To verify this, we visualize the distribution of CLIP scores of generated images in Figure \ref{fig:stable_CLIP} for different categories in the LVIS dataset. We find that StableDiffusion achieves similar CLIP score distribution in generating rare categories, showing it does not suffer serious data imbalance issues and performs similarly well for rare categories.

\subsection{Instance Composition} 
When applying X-Paste to compose training images, we sample instances with a class-balanced sampling strategy and paste them at random locations of the background images. The instances are resized to proper scales based on the resolution of the background image and a scale factor (percentage of object area in the background image) depending on data distribution. The background images can be either images with annotations or plain background images, we only use the training data for versatility. When the instance occludes an object in the background image, we remove fully occluded objects and update mask and bounding box annotations accordingly.

\noindent\textbf{Class-balanced Sampling:} We randomly choose the number of instances $N_i \in [1, N_{max}]$ ($N_{max}=20$ by default) for each training image, then we sample $N_i$ categories and one instance for each category in a repeatable way.

\begin{figure*}
\centering
\includegraphics[width=0.95\linewidth]{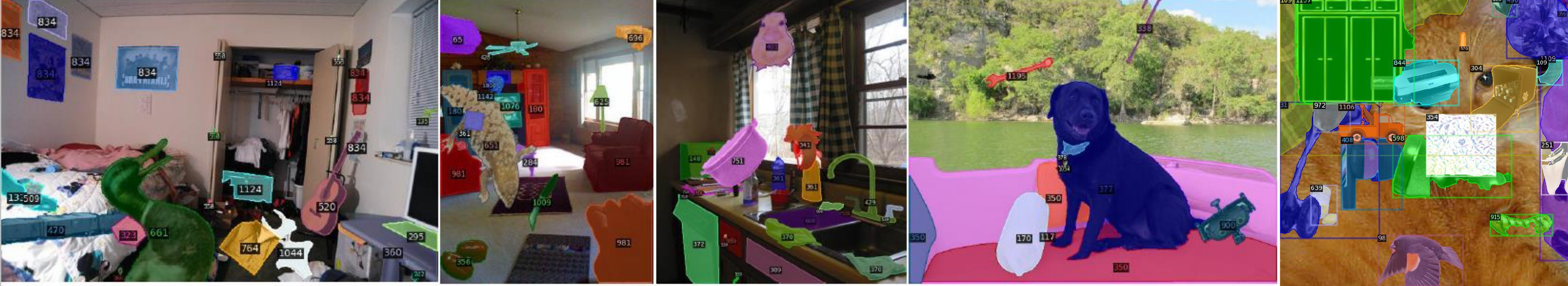}
\vspace{-1em}
\caption{Visualization of training samples synthesized by X-Paste.}
\label{fig:paste}
\vspace{-1em}
\end{figure*}

\noindent\textbf{Instance Composition:} We calculate the mean $\mu_{C}$ and standard variance $\sigma_{C}$ of object scale (square root of mask-area-divided-by-image-area) for each class $C$ in training set. When pasting instance $I$ with class $C$, we sample scale $S_r$ from Gaussian distribution $N(\mu_{C},\sigma^2_{C})$ and paste the instance with scale $S_r^2HW$ on background image ($H,W$ indicate the image height and width), where the coordinates of the center point of the bounding box are randomly selected within the whole image. In Figure \ref{fig:paste}, we show some training samples generated by X-Paste. These generated images are different from synthesized images with the previous methods~\cite{Instaboost, Contextual-Copy-Paste}, which aims to generate realistic images for training. Our generated images are not real since the  scale and relative relationship of the objects may not be reasonable. Still, these generated images can be used as good training data for improving performance.

\section{Experiments}
\label{sec:exp}

\subsection{Settings}
\noindent \textbf{Datasets.} We conduct experiments of object detection and instance segmentation on LVIS~\cite{LVIS} and MS-COCO~\cite{COCO} datasets. LVIS dataset contains 100k training images, and 20k validation images. It has 1203 categories with a long-tailed distribution of instances in each category. These categories can be  divided into common, frequent, and rare categories according to the instances number of each category. And the number of the common, frequent, and rare categories is 461, 405, and 337. In the open-vocabulary instance detection setting, common and frequent categories are for training and rare categories serve as the novel categories for testing. MS-COCO dataset contains 118K training, 5K validation, and 20K test-dev images. We use the official split for training. 

\noindent \textbf{Learning Framework.} We utilize the typical object detection and instance segmentation framework CenterNet2~\cite{CenterNet2} in Detectron2~\cite{wu2019detectron2}. Our basic setting uses Resnet 50~\cite{Resnet} pretrained on ImageNet-22k as backbone. The training configurations are set as follows: training resolution is set to 640, the batch size is 32, and 4$\times$ schedule (48 epochs). We test the performance with a single-scale strategy. We report the box AP and mask AP on all categories(donated as $AP^{box}$ and $AP^{mask}$) as well as on the rate categories(donated as $AP_r^{box}$ and $AP_r^{mask}$). 


\noindent \textbf{Baseline Settings of X-Paste.} We design a baseline setting to ablate X-Paste. To make sure each category contains enough instances, we use the CLIP model~\cite{CLIP} to filter 1k real images per category. Besides, we apply the StableDiffusion model~\cite{ldm} to generate 1k images for each category. For StableDiffusion, the diffusion steps is set to 200 with the classifier-free scale set to 5.0. We adopt the methods(donated as \emph{max CLIP}) described in Sec~\ref{sec:instance_mask_generation} as the default setting of instance mask generation. For Instance Filtering, we set the CLIP threshold as 0.21 to filter all the obtained instances. We keep 150k generated instances from StableDiffusion and 150k real instances from the CLIP model. During the Instance Composition module, the number of instances  pasted to each background image is set to 20 for training.

\subsection{Analysis of X-Paste}
\label{sec:instance}
We perform ablation studies to analyze our proposed X-Paste and find several intriguing properties.

\begin{table}[t]
  \small
  \centering
  \resizebox{\linewidth}{!}{
      \begin{tabular}{c|c|cccc}
      \toprule
      \# Gen & \# Real&$AP^{box}$&$AP^{mask}$&$AP_r^{box}$&$AP_r^{mask}$ \\
      \midrule
      0 & 0  & 34.5&30.8 & 24.0&21.6 \\
      100k & 100k & 36.3&32.3  &  28.0&25.2\\
      150k & 150k & 36.6&32.7 & 28.5&26.5  \\
      300k & 300k & \textbf{36.7}&\textbf{32.9} & \textbf{29.9}&\textbf{27.6} \\
    \bottomrule
    \end{tabular}
    }
\vspace{-2mm}
\caption{Results of using different scales of filtered instances in X-Paste. \# Gen and \# Real denotes the number of generated images from StableDiffusion and real images filtered by the CLIP model.}
\vspace{-2mm}
\label{tab:aba_gen_scale}
\end{table}

\begin{table}[t]
  \small
  \centering
  \resizebox{\linewidth}{!}{
      \begin{tabular}{c|c|cccc}
      \toprule
      \# Gen & \# Real&$AP^{box}$&$AP^{mask}$& $AP_r^{box}$&$AP_r^{mask}$\\
      \midrule
      300k  & 0 & 35.3&31.6 & 26.3&24.5 \\
      0 & 300k  & 36.1&32.0 & 27.8&25.0 \\
      150k & 150k & \textbf{36.6} & \textbf{32.7} & \textbf{28.5} & \textbf{26.5} \\
    \bottomrule
    \end{tabular}
    }
\vspace{-2mm}
\caption{Comparison of using different instances  sources in X-Paste. \# Gen and \# Real denotes the number of generated images from StableDiffusion and real images filtered by the CLIP model.}
\vspace{-2mm}
\label{tab:aba_gen_sources}
\end{table}

\noindent \textbf{More instances, better results.} In Table~\ref{tab:aba_gen_scale}, we compare the results without using and using different numbers of instances in X-Paste. We find that using only 100k generated instances or filtered real instances from X-Paste can boost performance significantly. Concretely, it achieves 36.3 box AP and 32.3 mask AP, surpassing baseline by +1.9 mAP and +1.5 mAP. It achieves a more significant gain on rare categories with a +4.0 box AP and +3.6 mask AP. More importantly, if we keep increasing the scale of instances from 100k to 300k, we can obtain better results. This also demonstrates the high value of using text-to-image generated images, as retrieving a large quantity of real object-centric images from search engine is very challenging. To further show this point, we use all the one million retrieved images on LVIS dataset, and compare the results with the baseline of using one million retrieved images + one million generated images in table \ref{tab:upperbound}. The results show that, even with a large amount of retrieved images, adding text-to-image generated images can still improve $0.7\sim0.9$ on mask AP.
\begin{table}[h]
  \small
  \centering
  \resizebox{\linewidth}{!}{
      \begin{tabular}{c|c|cccc}
      \toprule
      Backbone&Data&$AP^{box}$&$AP^{mask}$&$AP_r^{box}$&$AP_r^{mask}$ \\
      \midrule
      ResNet50 & Retr & 36.5 & 32.3 & 28.3 & 25.2 \\
      ResNet50 & Retr+Gen & 36.7 & 33.0 & 29.6 &27.8 \\
      Swin-L & Retr & 48.9 & 43.5 & 47.6 & 41.6 \\
      Swin-L & Retr+Gen & 50.1& 44.4 & 48.2 & 43.3 \\
    \bottomrule
    \end{tabular}
    }
\caption{Comparison with directly use generated images as training data.}
\label{tab:upperbound}
\end{table}

\noindent \textbf{Diverse instance sources matter.} We present the results of using different sources of instances for X-Paste in Table~\ref{tab:aba_gen_sources}. Under the same instances scale, Using CLIP filtered real images could achieve better results. Moreover, we observe that using both instances from the generative model and filtered real images can achieve a better performance than using single instance sources. This validates that the diversity of these instances is of great importance for boosting performance.

\noindent \textbf{Effect of instance mask generation.}
An accurate instance mask serves an important role in training object detection and instance segmentation. Since an imprecise mask may lead the model to learn wrong knowledge. We study several category-agnostic segmentation methods here. Concretely, we experiment with the instance mask generated by U2Net\cite{U2Net} and SelfReformer\cite{selfreformer}, UFO\cite{UFO}, CLIPseg \cite{CLIPseg} our proposed method to select the mask segmentation with the max CLIP score. Table~\ref{tab:aba_seg} reports the results of using different instance mask generation methods. We find that these segmentation methods can achieve comparable results. Our proposed max CLIP could yield slightly better results in $AP_r^{box}$, $AP_r^{mask}$, $AP^{box}$, and $AP^{mask}$.

\begin{table}[t]
  \small
  \centering
  \resizebox{\linewidth}{!}{
      \begin{tabular}{c|cccc}
      \toprule
      Foreground segmentation&$AP^{box}$&$AP^{mask}$&$AP_r^{box}$&$AP_r^{mask}$ \\
      \midrule
      U2Net~\cite{U2Net} & 35.6 & 31.7  & 27.1 & 24.8 \\
      CLIPseg~\cite{CLIPseg} &  35.4 & 31.6 & 25.5 & 23.1  \\
      selfreformer~\cite{selfreformer} & 35.5 & 31.7 & 26.6 & 24.1 \\
      UFO~\cite{UFO}  & 35.7 & 31.7 & \textbf{27.5} & \textbf{25.0} \\
      max CLIP & \textbf{35.8} & \textbf{31.8} & 26.2 & 23.7 \\
    \bottomrule
    \end{tabular}
    }
\vspace{-2mm}
\caption{Comparison of using different segmentation methods.}
\vspace{-4mm}
\label{tab:aba_seg}
\end{table}

\begin{table}[t]
  \small
  \centering
  \resizebox{\linewidth}{!}{
      \begin{tabular}{c|cccc}
      \toprule
      CLIP score threshold&$AP^{box}$&$AP^{mask}$&$AP_r^{box}$&$AP_r^{mask}$ \\
      \midrule
      0  & 36.0 & 32.1 & 27.6 & 25.1 \\
      0.2 & 36.4 & 32.5 & \textbf{29.0} & 26.1 \\
      0.21 & \textbf{36.6} & \textbf{32.7} &28.5 & 26.5  \\
      0.22 & 36.4 & 32.4 & 28.3 & 25.7\\
      0.23 & 36.3 & 32.4 & 28.8 & \textbf{26.7} \\
      0.24 & 36.3 & 32.5 & 28.1 & 25.8  \\
      0.25 & 36.3 & 32.4 & 28.4 & 25.8\\
    \bottomrule
    \end{tabular}
    }
\vspace{-2mm}
\caption{Ablation of CLIP score thresholds for instance filtering.}
\label{tab:CLIP_thres}
\end{table}
 \begin{table}[t]
  \small
  \centering
  \resizebox{\linewidth}{!}{
  \begin{tabular}{l|ccccc}
  \toprule
  Placement & $N_{max}$ &$AP^{box}$&$AP^{mask}$ &  $AP_r^{box}$&$AP_r^{mask}$\\
  \midrule
  reference & - & 33.7 & 30.1 & 22.9 & 21.2 \\
  random & 10 & 36.2 & 32.2 &  28.3 & 25.1 \\
  random & 20 & \textbf{36.6} & \textbf{32.7} &\textbf{28.5} & \textbf{26.5 } \\
  random & 30 & 35.5 & 31.3 &26.8 & 24,6  \\
\bottomrule
\end{tabular}
}
\caption{ Ablation of different placement strategies. }
\label{tab:aba_place}
\end{table}

\begin{table}[t]
  \small
  \centering
  \resizebox{\linewidth}{!}{
    \begin{tabular}{l|cccc}
      \toprule
      Method &AP$^{box}$ & AP$^{mask}$ & AP$_r^{box}$ & AP$_r^{mask}$\\
      \midrule
      baseline & 34.5 & 30.8 & 24.0 & 21.6  \\
      baseline+External Data & 35.3 & 31.7 & 29.5 & 27.4 \\
      Copy-Paste\cite{Simple-Copy-Paste}  & 35.4 & 31.5 &  25.3 & 22.1\\
      Detic\cite{Detic} & 35.3 & 31.7 &  27.5 & 25.4 \\
      X-Paste  & \textbf{36.6} & \textbf{32.7} & \textbf{28.5} & \textbf{26.5}\\
    \bottomrule
    \end{tabular}
  }
\vspace{-2mm}
\caption{Comparison with related methods on LIVS dataset. We use the same network and training settings here. We use 150k generated images and 150k retrieved images to boost object detection and instance segmentation with X-Paste and Detic.}
\vspace{-4mm}
\label{tab:LVIS2}
\end{table}

\begin{table*}[t]
  \small
  \centering
  \resizebox{0.85\linewidth}{!}{
    \begin{tabular}{l|l|rrrr}
      \toprule
      Method& Backbone &AP$^{box}$&AP$^{mask}$&AP$_r^{box}$&AP$_r^{mask}$\\
      \midrule
      Copy-Paste\cite{Simple-Copy-Paste} &  EfficientNet-B7 &  41.6&38.1& -&32.1\\
      Tan et al.\cite{Tan20201stPS} & ResNeSt-269  & - & 41.5 & - & 30.0 \\ 
      Detic\cite{Detic} & Swin-B & 46.9&41.7 & 45.9 & 41.7 \\
      \midrule
      CenterNet2~\cite{CenterNet2} &  Swin-L & 47.5 & 42.3 & 41.4 & 36.8 \\
      w/ Copy-Paste\cite{Simple-Copy-Paste} &  Swin-L & 49.5 & 43.7 & 43.4 & 38.5 \\
      w/ X-Paste & Swin-L  & \textcolor{blue}{(+2.6)} \textbf{50.1} & \textcolor{blue}{(+2.1)} \textbf{44.4} & \textcolor{blue}{(+6.8)} \textbf{48.2} & \textcolor{blue}{(+6.5)} \textbf{43.3} \\
       \rowcolor{gray!30} w/ X-Paste + Copy-Paste & Swin-L & \textcolor{blue}{(+3.4)} \textbf{50.9} & \textcolor{blue}{(+3.1)} \textbf{45.4} & \textcolor{blue}{(+7.3)} \textbf{48.7} & \textcolor{blue}{(+7.0)} \textbf{43.8} \\
    \bottomrule
    \end{tabular}
  }
\caption{Comparision with previous methods on LVIS 1.0 validation set. }
\label{tab:LVIS1}
\end{table*}

\noindent \textbf{Effect of using different CLIP score thresholds for instance filtering. }
With different CLIP score thresholds, we choose a fixed number of instances for training the framework. We report the results in Table~\ref{tab:CLIP_thres}. we find 0.21 is the best threshold, comparing non-filtering (threshold set to 0), it provides a gain of 0.6 mask AP for all classes and 1.4 mask AP for rare classes. We also observe that a higher CLIP score threshold may not produce better performance. We believe the reason is that higher CLIP scores mean the generated samples are more correlated with given texts but they may show limited diversity in terms of appearance, viewpoint, and style. The lack of diversity will cause worse performance. This proves again that both quality and diversity of instances are of great importance for training instance segmentation.

\noindent \textbf{Study different instances placement strategies.} We compare our random placement strategy with a reference-based placement strategy. Reference-based placement strategy pastes the generated instances to the location of the original bounding box in the background image. We also study how the maximum number of instances $N_{max}$ pasted a single background image affects the performance. Table~\ref{tab:aba_place} reports the results. We observe random placement strategy achieves a better performance than reference-based methods. Besides, we find that 20 is a suitable number for max number of instances pasted to a background image.

\subsection{Comparison with the previous methods}
\label{sec:comp}
\noindent \textbf{Comparison with other data-augmentation methods.}
In table \ref{tab:LVIS2}, we make a fair comparison with previous data-augmentation related methods. We use the same framework CenterNet2~\cite{CenterNet2} with Resnet50 as the backbone. We compare our method with Copy-Paste \cite{Simple-Copy-Paste} and Detic\cite{Detic} on LVIS dataset. Table~\ref{tab:LVIS2} reports the results on LVIS dataset. As the reference, we also show another baseline (``baseline+External data") that directly adds our collected data into the training data without augmentation. We train Detic with the same instances as X-Paste to investigate whether X-Paste is more efficient than weakly-supervised learning with image-level annotations. Compared with the strong baseline Copy-Paste~\cite{Simple-Copy-Paste}, our method achieves significantly better results. In all categories, X-Paste outperforms Copy-Paste by +1.2 box AP and +1.2 mask AP. In the rare categories, X-Paste achieves a larger gain with +3.2 box AP and +4.4 mask AP. This validates the effectiveness of a large scale of instances is crucial for achieving better results. Our X-Paste also outperforms the recently proposed Detic~\cite{Detic}. Compared with Detic, X-Paste does not involve a complicated mechanism and not requires a larger batch size. This validates the effectiveness of the proposed X-Paste.

To verify whether X-Paste is suitable for large models. We further conduct experiments with the recently proposed large model Swin-Large~\cite{Swin}(Swin-L for short) as the backbone. We increase the input resolution to 896 and train the model for 4 $\times$ schedule (~48 epochs) with exponential moving average to update the model. 
We use about 1 million generated and retrieved instances filter with the CLIP score threshold set to 0.21 for X-Paste. 
Table~\ref{tab:LVIS1} lists the results. We observe that X-Paste outperforms Copy-Paste by +0.6 box AP and +0.7 mask AP in all categories, and +4.8 box AP and +4.8 mask AP in rare categories. X-Paste can provide more improvements on rare categories since X-Paste offers many instances for rare categories while Copy-Paste can not. This validates that X-Paste could achieve consistent gain over Copy-Paste on top of solid baselines. More importantly, we can combine X-Paste with Copy-Paste to have more instances for pasting to background images for training. We observe a more strong performance. This further validates that the diversity of instances is important.

\noindent \textbf{Open-Vocabulary Object Detection.}  We follow the setting of previous work and remove the images containing rare categories from the training set. Then we train the framework on the images which contain common and frequent categories. We adopt CLIP classifier\cite{ViLD} and initialize our model with the Box-Supervised baseline from Detic \cite{Detic}. In Table \ref{tab:openLVIS}, we evaluate the open vocabulary detection performance of different methods on LVIS dataset. Compared with CLIP based methods like ViLD~\cite{ViLD}, RegionCLIP~\cite{RegionCLIP}, PromptDet~\cite{PromptDet}, and DetPro~\cite{DetPro}. X-Paste achieves a significantly better performance. This shows that, without any architecture or algorithm change, the simple Copy-Paste by using generated or filtered novel instances is already an effective and strong baseline for open-vocabulary object detection. We also notice that previous methods like Detic~\cite{Detic}, MEDet~\cite{MEDet} and Centirc-OVD~\cite{Centric-OVD} achieve better performance, they use external human-annotated data like ImageNet-22k and need to make big architecture or algorithm change accordingly to consume such data. It is worthy to note that, as an simple data augmentation plugin, our X-Paste is orthogonal and complementary with these dedicated designed methods.

\begin{table}[t]
  \small
  \centering
  \resizebox{\linewidth}{!}{
  \begin{tabular}{l|cccc}
  \toprule
  Method &AP$^{box}$&AP$^{mask}$&AP$_{novel}^{box}$&AP$_{novel}^{mask}$\\
  \midrule
  ViLD\cite{ViLD} & 27.5 & 25.2 & 17.4 & 16.8 \\
  RegionCLIP\cite{RegionCLIP} & - & 28.2 & - & 17.1 \\
  PromptDet\cite{PromptDet} & - & 25.5 & - & 21.7 \\
  DetPro\cite{DetPro} & 28.4 & 25.9 & 20.8 & 19.8  \\
  Box-Supervised\cite{Detic} & 33.8 & 30.2 & 17.6 & 16.4 \\
  \midrule
  X-Paste & 35.7 & 31.8 & 22.8 & 21.4 \\

 \midrule
  Detic$^*$\cite{Detic} & 36.3 & 32.4 & 26.7 & 24.9 \\
  MEDet$^*$\cite{MEDet} & - & 34.4 & - & 22.4 \\
  Centric-OVD$^*$\cite{Centric-OVD} & - & 32.9 & - & 25.2 \\
\bottomrule
\end{tabular}
}
\caption{Comparison of open-vocabulary detection performance on LVIS, $^*$ means they use external supervised data for training.}
\label{tab:openLVIS}
\end{table}

\noindent \textbf{Performance on COCO dataset.} We also conduct experiments on the COCO dataset with our proposed X-Paste. The COCO dataset is more challenging since it has a large number of instances for each category, which does suffer from the long-tailed distribution issue. We apply the StableDiffusion model with the same setting as on LVIS to generate 5k images for each category. For Instance Filtering, we set the CLIP threshold as 0.21 to filter all the obtained instances. We keep 300k generated instances from StableDiffusion and 80k real instances from the CLIP model. During the Instance Composition module, we paste 20 instances to each background image for training as well. We conduct experiments on the CenterNet2 framework with two backbones: ResNet50 and Swin-L. We report the results Table~\ref{tab:COCO1}. With ResNet50 backbone, X-Paste outperforms the baseline by 2.5 box AP and 1.8 mask AP. With Swin-L backbone, X-Paste achieves 57.0 box AP and 48.5 mask AP, surpassing baseline by +1.7 box AP and 0.9 mask AP. We can find that our methods could consistently improve the baseline in various settings on the COCO dataset. By combining X-Paste with Copy-Paste, we observe a more strong performance as well. Even for the COCO dataset with multiple instances in each category, X-Paste is still effective.

\begin{table}[t]
  \small
  \centering
  \resizebox{\linewidth}{!}{
      \begin{tabular}{l|l|r r}
      \toprule
      Method& Backbone &AP$^{box}$ & AP$^{mask}$\\
      \midrule
      CenterNet2~\cite{CenterNet2} & ResNet50 & 46.0 & 39.8 \\
      w/ Copy-Paste\cite{Copy-Paste} & ResNet50 & 46.4 & 39.8 \\
      w/ X-Paste & ResNet50 & \textcolor{blue}{(+0.6)} 46.6&
      \textcolor{blue}{(+0.1)} 39.9 \\
      w/ X-Paste + Copy-Paste & ResNet50 &  \textcolor{blue}{(+0.8)} 46.8 & \textcolor{blue}{(+0.2)} 40.0 \\
      \midrule
      CenterNet2~\cite{CenterNet2} & Swin-L & 55.3&47.7 \\
      w/ Copy-Paste\cite{Copy-Paste} & Swin-L & 56.1 & 48.3 \\
      w/ X-Paste & Swin-L & \textcolor{blue}{(+1.3)} 56.6 & \textcolor{blue}{(+0.9)} 48.6 \\
      w/ X-Paste + retrieval & Swin-L & \textcolor{blue}{(+1.5)} 56.8 & \textcolor{blue}{(+1.0)} 48.7 \\
      w/ X-Paste + Copy-Paste & Swin-L & \textcolor{blue}{(+1.6)} 56.9 & \textcolor{blue}{(+1.1)} 48.8 \\
    \bottomrule
    \end{tabular}
}
\caption{X-paste works well across a variety of different model architectures on the COCO dataset.}
\label{tab:COCO1}
\end{table}

\section{Conclusion}
This paper revisits the previous Copy-Paste data augmentation methods and finds it is effective while showing limited scale-up capability. To this end, we propose a scalable version of Copy-Paste called X-Paste. X-Paste leverages the zero-shot recognition models(\eg CLIP) and text2image models(\eg StableDiffusion) to obtain large-scale images with accurate categories. Then these images are further transformed into instances through a series of carefully designed modules. Finally, these instances can be used in data augmentation for training instance segmentation. X-Paste provides significant improvements on top of strong baselines on LVIS and COCO datasets. We hope that X-Paste will foster further research on utilizing zero-shot recognition or generative models for various vision tasks.

\section{Acknowledgements}
This work was partially supported in part by the Fundamental Research Funds for the Central Universities under Grant WK5290000003, the Natural Science Foundation of China under Grant U20B2047, 62121002, 62072421, 62002334 and Key Research and Development program of Anhui Province under Grant 2022k07020008. 

\bibliography{example_paper}
\bibliographystyle{icml2023}

\end{document}